\documentclass[11pt]{article}
\usepackage{lipsum}
\usepackage{background}
\backgroundsetup{
  position=current page.east,
  angle=-90,
  nodeanchor=east,
  vshift=-5mm,
  opacity=1,
  scale=3,
  contents=Draft
}

\usepackage{eacl2017}
\usepackage{times}
\usepackage{url}
\usepackage{latexsym}
\usepackage{mathtools}
\usepackage{graphicx}
\usepackage{svg}
\usepackage{subfig}
\usepackage{tikz}
\usepackage{tikz-dependency}
\usetikzlibrary{arrows,positioning}
\usepackage{amsmath}

\eaclfinalcopy % Uncomment this line for the final submission

\title{Universal Dependencies to Logical Forms with Negation Scope}

\author{
Federico Fancellu\ \ \ \ \ \ Siva Reddy\ \ \ \ \ \ Adam Lopez\ \ \ \ \ \  Bonnie Webber\\
ILCC, School of Informatics, University of Edinburgh \\
{\tt\small f.fancellu@sms.ed.ac.uk, siva.reddy@ed.ac.uk, \{alopez, bonnie\}@inf.ed.ac.uk}\\
}

\date{}

\begin{document}
\maketitle
\begin{abstract}
Many language technology applications would benefit from the ability to represent negation and its scope on top of widely-used linguistic resources. In this paper, we investigate the possibility of obtaining a first-order logic representation with negation scope marked using \textit{Universal Dependencies}. To do so, we enhance \textit{UDepLambda}, a framework that converts dependency graphs to logical forms. The resulting \textit{UDepLambda$\lnot$} is able to handle phenomena related to scope by means of an higher-order type theory, relevant not only to negation but also to universal quantification and other complex semantic phenomena. The initial conversion we did for English is promising, in that one can represent the scope of negation also in the presence of more complex phenomena such as universal quantifiers.
\end{abstract} 

\section{Introduction}
Amongst the different challenges around the topic of negation, detecting and representing its scope is one that has been extensively researched in different sub-fields of NLP (e.g. Information Extraction \cite{velldal2012speculation,fancellu2016neural}). In particular, recent work have acknowledged the value of representing the scope of negation on top of existing linguistic resources (e.g. AMR -- \newcite{bos2016expressive}). Manually annotating the scope of negation is however a time-consuming process, requiring annotators to have some expertise of formal semantics.\\
\indent Our solution to this problem is to automatically convert an available representation that captures negation into a framework that allows a rich variety of semantic phenomena to be represented, including scope. That is, given an input sentence, we show how its \textit{universal dependency} (UD) parse can be converted into a representation in first-order logic (FOL) with lambda terms that captures both predicate--argument relations and scope.\\
\begin{figure}[t]
\centering
\subfloat[UD Dependency Tree]{
\begin{dependency}[theme = simple]
   \begin{deptext}[column sep=1em]
      Malta \& borders \& no \& country\\\\
   \end{deptext}
   \deproot{2}{ROOT}
   \depedge{2}{1}{NUSBJ}
   \depedge[arc angle=100]{4}{3}{NEG}
   \depedge{2}{4}{DOBJ}
\end{dependency}
}\\
\subfloat[UDepLambda Logical Form]{\shortstack{$\lambda e. \exists x\exists y.borders(e) \land country(x) \land no(x) \land $\\$ Malta(y) \land arg1(e,y) \land arg2(e,x)$}}\\
\subfloat[Desired Logical Form]{\shortstack{$\forall x. country(x) \to \lnot \exists e\exists y. borders(e) \land $\\$ Malta(y) \land arg1(e,y) \land arg2(e,x)$}}\\
\caption{The dependency tree for \textsl{`Malta borders no country'} and its logical forms}
\label{introex}
\end{figure}
\indent Our approach is based on \textit{UDepLambda} \cite{reddy2017universal,reddy2016transforming}, a constraint framework that converts dependency graphs into logical forms, by reducing  the lambda expressions assigned to the dependency edges using the lambda expressions of the connected head and child nodes. 
The edge labels in the input UD graph are only edited minimally so to yield a more fine-grained description on the phenomena they describe, while lexical information is used only for a very restricted class of lexical items, such as negation cues. A FOL representation of the entire input graph can be then obtained by traversing the edges in a given order and combining their semantics.\\
\indent However, in its original formulation, UDepLambda does not handle either universal quantifiers or other scope phenomena. For example, the sentence `Malta borders no country' has the UD graph shown in Figure~\ref{introex}(a). When compared to the correct representation given in Figure~\ref{introex}(c), the UDepLambda output shown in Figure~\ref{introex}(b) shows the absence of universal quantification, which in turn leads negation scope to be misrepresented.\\ 
\indent For this reason, we set the foundation of {\bf UDepLambda$\lnot$} (UDepLambda-\textit{not}), an enhanced version of the original framework, whose type theory allows us to jointly handle negation and universal quantification. Moreover, unlike its predecessor, the logical forms are based on the one used in the `Groeningen Meaning Bank' (GMB; \cite{basile2012developing}), so to allow future comparison to a manually annotated semantic bank.\\
\indent Although the present work shows the conversion process for English, given that the edge labels are \emph{universal}, our framework could be used to explore the problem of representing the scope of negation in the other 40+ languages universal dependencies are available in. This could also address the problem that all existing resources to represent negation scope as a logical form are limited to English (e.g. GMB and `DeepBank' \cite{flickinger2012deepbank}) or only to a few other languages (e.g. `The Spanish Resource Grammar' \cite{marimon2010spanish}).\\
\indent In the reminder of this paper, after introducing the formalism we will be working in (\S \ref{depplus}), we will work the theory behind some of the conversion rules, from basic verbal negation to some of the more complex phenomena related to negation scope, such as the determiner `no'(\S \ref{no}), the interaction between the negation operator and the universal classifier (\S \ref{univclass}) and non-adverbial or lexicalized negation cues such as `nobody', `nothing' and `nowhere' (\S \ref{nonadv}). Limitations, where present, will be highlighted.\\
\indent {\bf Contribution}. The main contribution of the paper is \textit{UDepLambda$\lnot$}, a UD-to-FOL conversion framework, whose type theory is able to handle scope related phenomena, which we show here in the case of negation.\\
\indent {\bf Future work}. \textit{UDepLambda$\lnot$} can serve as a basis for further extensions that could apply to other complex semantic phenomena and be learned automatically, given the link to a manually annotated semantic bank.

\section{UDepLambda$\lnot$} \label{depplus}
\begin{figure}
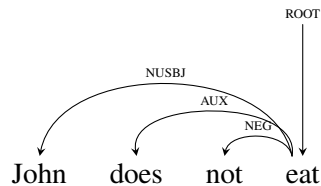

\centering
\begin{dependency}[theme = simple]
   \begin{deptext}[column sep=1em]
      John \& does \& not \& eat\\\\
   \end{deptext}
   \deproot{4}{ROOT}
   \depedge{4}{1}{NUSBJ}
   \depedge[arc angle=100]{4}{2}{AUX}
   \depedge{4}{3}{NEG}
\end{dependency}
\caption{Dependency graph for the sentence 'John does not eat'}
\label{ex1}
\end{figure}

We introduce here the foundations of \textit{UDepLambda$\lnot$}, an enhancement to the \textit{UDepLambda} framework to convert a UD graph into its correspondent logical form. As its predecessor, the conversion takes place in four different steps: \emph{enhancement}, \emph{binarization}, \emph{substitution} and \emph{composition}. Whereas binarization and composition are the same as \textit{UDepLambda}, substitution differs in:
\begin{itemize}
\item  using a higher order type-theory to deal with universal quantification, which can interact with other scope operator such as negation;
\item using FOL expressions based on those used in the Groeningen Meaning Bank (GMB), so as to link to a manually--annotated semantic bank which can be leveraged for future work.\footnote{The current study ignores certain aspects of Discourse Representation Theory \cite{kamp2011discourse} on which the GMB is based, which are secondary to the issues we are focussed on.}
\end{itemize}
The details of the four steps are as follows:\\

{\bf Enhancement}. In this step, we first convert a dependency tree to a dependency graph using existing existing enhancements in UDepLambda.
The enhanced dependency labels are represented in red color.
In future, we will replace this step with existing enhancements \cite{schuster2016enhanced}.

{\bf Binarization}. The dependency graph is mapped to a LISP-style s-expression, where the order of the edge traversal is specified. For instance, the expression (\textit{nsubj} (\textit{aux} (\textit{neg} eat not) does) John) indicates that the semantic representation of the sentence in Figure (\ref{ex1}) is derived by composing the semantics of the edge \textit{nsubj} with the logic form of `John' and of the phrase 'does not eat'. The semantics of the phrase `does not eat' is in turn derived by composing the edge \textit{aux} with the phrase `not eat' and the auxiliary `does'. Finally `not' and `eat' are composed along the edge \textit{neg}.\\
\indent The order of traversal follows an \textit{obliqueness hierarchy} which defines a strict ordering of the modifiers of a given head traversed during composition. This hierarchy is reminiscent of bottom-up traversal in a binarized constituency tree (where for instance the direct object is always visited before the subject). Furthermore, for a head to be further composed, all its modifiers needs to be composed first. In the sentence in Figure (\ref{ex1}), this hierarchy is defined as \textit{neg} $>$ \textit{aux} $>$ \textit{nusbj}, where the semantics of the subject can be applied only when the other modifiers to the verb-head have been already composed.\\
\indent {\bf Substitution}. The substitution step assigns a lambda expression to each edge and vertex (i.e. word) in the graph. \textit{The lambda expressions of the edges are manually crafted to match the semantics of the edge labels while no assumption is made on the semantics of the word-vertices which are always introduced as existentially bound variables}. This allows us\textit{not} to rely for most part on any language-specific lexical information. These expressions follows recent work on semantic compositionality of complex phenomena in event semantics \cite{champollion2011quantification}. In doing this, we generalize our type theory as follows:
\begin{itemize}
\item Each word-vertex is assigned a semantic type $\langle \langle v,t \rangle,t\rangle$ or $\langle \langle v,t \rangle,t\rangle$ (here shortened in $\langle vt,t \rangle$), where $v$ stands for either a paired variable of type $\mathrm{Event} \times \mathrm{Individual}$. This is in contrast with the type assigned to words in the original UDepLambda $\langle v,t \rangle$. The result of this type-raising operation is clear when we compare the following lambda expressions:\\\\
UDepLambda: $\lambda x. man(x_a)$\\
UDepLambda$\lnot$: $\lambda f. \exists x. man(x_a) \land f(x)$\\\\
where the `handle' $f$ allows for complex types to be added inside another lambda expression.\\
Following the GMB, proper nouns are treated like indefinite nouns, being linked to a existentially-bound variable (e.g. John := $\lambda f.\exists x. named(x_a,John,PER) \land f(x)$).
\item Each edge is assigned the semantic type $\langle\langle vt,t\rangle, \langle\langle vt,t\rangle, \langle vt,t\rangle\rangle\rangle$ where we combine a generalized quantifier over the parent word (P) with the one over the child word (Q) to return another generalized quantifier (f). For instance, when reducing the sub-expression (\textit{nsubj} eat John), we first reduce the parent vertex `eat' (P) and then the child vertex `John'(Q) using the semantics of the subject (`Actor' in the GMB).\\\\
nsubj:= $\lambda P.\lambda Q.\lambda f. P(\lambda x. f(x) \land Q(\lambda y.$\\$   Actor(x_e, y_a)))$\\\\
When compared to the original UDepLambda expression (of type $\langle\langle v,t\rangle, \langle\langle v,t\rangle, \langle v,t\rangle\rangle\rangle$):\\\\
 $\lambda f.\lambda g. \lambda x.\exists y. f(x_e) \land g(y_a) \land arg1(x_e,y_a)$\\\\
unlike its predecessor, UDepLambda$\lnot$ allows for nested dependencies between parent and child node which is necessary to model scope phenomena.
\item In cases such as the sub-expression (\textit{neg} `John does eat' not), the edge label \textit{neg} and the word `not' carry the exact same semantics (i.e. the negation operator $\lnot$). For these \textit{functional words} we try to define semantics on the dependency edges only rather than on the word. As shown below,  reducing Q does not impact the semantic composition of the edge \textit{neg}:
\begin{center}
{\small neg:= $\lambda P.\lambda Q.\lambda f. \lnot P(\lambda x.f(x))$}\\
{\small not:= $\lambda f.TRUE$}
\end{center}
\end{itemize}
\indent {\bf Composition}. The lambda expressions are reduced by following the traversal order decided during the \emph{binarization} step. Let's exemplify the composition step by showing at the same time how {\bf simple verbal negation} composes semantically, where the input s-expression is (\textit{neg} (\textit{aux} (\textit{nsubj} eat John) does) not). The substitution step assigns vertices and edges the following semantics:\\\\
`eat' := $\lambda f.\exists x. eat(x_e) \land f(x)$\\
`not' := $\lambda f.TRUE$\\
`John' := $\lambda f.\exists x. named(x_a,John,PER) \land f(x)$\\
`does' := $\lambda f.TRUE$\\\\
\textit{nsubj}:= {\small $\lambda P.\lambda Q.\lambda f. P(\lambda x. f(x)  \land Q(\lambda y. Actor(x_e, y_a)))$}\\
\textit{aux} := {\small $\lambda P.\lambda Q.\lambda f. P(\lambda x.f(x))$}\\
\textit{neg}:= {\small $\lambda P.\lambda Q.\lambda f. \lnot P(\lambda x.f(x))$}\\
\textit{ex-closure}:= $\lambda x.TRUE$\\\\
where the subscripts $e$ and $a$ stands for the event-type and the individual-type existential variable respectively. As for the edge \textit{neg}, the child of a \textit{aux} edge is ignored because not contributing to the overall semantics of the sentence.\footnote{The present work does not consider the semantics of time the word `does' might contribute to.}
We start by reducing (\textit{neg} eat not), where P is the parent vertex `eat' and Q the child vertex `not'. This yields the expression:\footnote{Step-by-step derivations are shown in Appendix A.}
\begin{center}
$\lambda f. \lnot \exists x. eat(x_e) \land f(x)$
\end{center}
We then use this logic form to first reduce the lambda expression on the edge \textit{aux}, which outputs the same input representation, and then compose this with the semantics of the edge \textit{nsubj}. The final representation of the sentence (after we apply existential closure) is as follows:
\begin{center}
$\lnot \exists x. \exists y. eat(x_e) \land named(y_a,John,PER) \land Actor(x_e,y_a)$
\end{center}
Given the resulting logical form we consider as part of negation scope all the material under the negation operator $\lnot$.

\section{Analysis of negative constructions}

\subsection{The quantifier `no'} \label{no}

\begin{figure}[t]
\begin{center}
\subfloat[Original UD Dependency Tree]{
\begin{dependency}[theme = simple]
   \begin{deptext}[column sep=1em]
      No \& man \& came.\\\\
   \end{deptext}
   \deproot{3}{ROOT}
   \depedge{3}{2}{NUSBJ}
   \depedge{2}{1}{NEG}
\end{dependency}
}\\
\subfloat[UDepLambda Logical Form]{\shortstack{$\exists x. came(x_e) \land \lnot \exists y. man(y_a) \land arg1(x_e,y_a)$}}\\
\subfloat[Enhanced UD Dependency Tree]{
\begin{dependency}[theme = simple]
   \begin{deptext}[column sep=1em]
      No \& man \& came.\\\\
   \end{deptext}
   \deproot{3}{ROOT}
   \depedge{3}{2}{\textcolor{red}{NUSBJ:INV}}
   \depedge{2}{1}{\textcolor{red}{NEG:UNIV}}
\end{dependency}
}\\
\subfloat[Desired Logical Form]{\shortstack{$\forall y. man(y_a) \to \lnot \exists x. came(x_e) \land Actor(x_e,y_a)$}}
\caption{The dependency trees for \textsl{`No man came'} (top: original UD tree; bottom: enhanced UD tree) and its logical forms}
\label{nomancame}
\end{center}
\end{figure}
Let's consider the sentence `No man came' along with its dependency trees and logical form, shown in Figure \ref{nomancame}.\\
\indent As shown in Figure~\ref{nomancame}(b), one shortcoming of the original UDepLambda is that it doesn't cover universal quantification. However, even if we were to assign any of the following lambda expressions containing material implication to the \emph{neg} edge connecting parent-$\lambda f$ (`man') and child-$\lambda g$ (`no'):
\begin{center}
?$\lambda f.\lambda g.\lambda x. f(x) \to \lnot f(x)$\\
?$\lambda f.\lambda g.\lambda x. f(x) \to g(x)$
\end{center}
the resulting expressions would have no means of later accommodating the event `came' in the consequent of the material implication:
\begin{center}
*$\lambda x. man(x) \to \lnot man(x)$\\
*$\lambda x. man(x) \to no(x)$
\end{center}
\indent The higher-order type theory of UDepLambda$\lnot$ solves this problem by ensuring that a) there is a universal quantified variable along with material implication and b) the entity bound to it ($man(x)$) is introduced only in the antecedent, whereas the negated event (along with other arguments) only in the consequent. The lambda expression assigned to the \textit{neg} edge is the following\\\\
$\lambda P.\lambda Q.\lambda f. \forall x.(P(\lambda y. EQ(x,y)) \to \lnot f(x))$\\\\
where $f$ allows to leave a `handle' for the event `came' to be further composed in the subsequent only, whereas the two-place function EQ($x$,$y$) as argument of $P$ binds the word in the parent node with the universally quantified variable $x$.\\
\indent It is worth mentioning at this point that although the universal quantifier `no' is parsed as depending from an edge \emph{neg}, it possesses a semantics that distinguishes it from other negative adverbs such as `not' or `never', in the fact that they bind their head to a universally quantifiable variable. In these cases we also \textit{enhance} the label on the dependency edge to reflect this more fine-grained distinction. In the presence of `no' the \emph{neg} edge becomes {\bf neg:univ} if its child vertex is a universal quantifier. This edit operation relies on having a list of lexical items for both universal quantifiers and negation cues in a language, which is easily obtainable given that these items form a small, closed class.\\
\indent A further edit operation is needed to make sure that the quantifier always outscopes the negation operator; to do so, we modify the semantics of the edge that connects the head of the edge \emph{neg:univ} (`man') with its parent (`came'), \emph{nsubj}, by inverting the order of the Q and P, so that the former outscopes the latter. We call this enhanced edge an `\emph{edge-name:}inv.' edge. Compared to \emph{nsubj}, the semantics of {\bf nsubj:inv} would be as follows:
\begin{center}
\emph{nsubj} := $\lambda P.\lambda Q.\lambda f. P(\lambda x. f(x) \land Q(\lambda y.  Actor(x_e,y_a)))$\\
\emph{nsubj-inv} := $\lambda P.\lambda Q.\lambda f. Q(\lambda y. P(\lambda x. Actor(x_e,y_a) \land f(x)))$
\end{center}
Using the edited input UD graph, the hierarchy we follow during composition is \textit{neg:univ} $>$ \textit{nsubj:inv} to yield the s-expression (\textit{nsubj:inv} (\textit{neg:univ} no man) came). Given the following input semantics:
\begin{center}
man:= $\lambda f. \exists x.man(x_a) \land f(x)$\\
came:= $\lambda f. \exists x.came(x_e) \land f(x)$\\
\emph{neg:univ}:= $\lambda P.\lambda Q.\lambda f. \forall x.(P (\lambda y. EQ(x,y)) \to \lnot f(x))$\\
\emph{nsubj:inv} := $\lambda P.\lambda Q.\lambda f. Q(\lambda y. P(\lambda x.Actor(x_e,y_a) \land f(x)))$
\end{center}
we first reduce the lambda expression on the edge \textit{neg:univ.} to yield the expression $\lambda f.\forall x.(man(x_a) \to \lnot f(x))$ and then combine it along the edge \textit{nsubj:inv} to yield the following representation:
\begin{center}
$\forall y.(man(y_a) \to \lnot \exists x. came(x_e)\land Actor(x_e,y_a))$
\end{center}
, where the scope of negation is correctly converted as inside the universal quantifier.\\
\indent Inverting the order of the parent and child nodes in the semantics of the \textit{:inv.} edge always allows to represent the universally quantified element as outscoping the event it depends on. At the same time, all other arguments and modifiers of the parent event will always compose inside the consequent. This applies to our initial example in Figure \ref{introex}, where composing the s-expression (\textit{dobj:inv.} borders `no country') to yield the expression:
\begin{center}
$\lambda f.\forall y.(country(y_a) \to \lnot \exists x. borders(x_e)\land Theme(x_e,y_a) \land f(x))$
\end{center}
, makes sure that further material can only be added in place of $f(e)$, which is inside the scope of $\lnot$, in turn in the scope of $\forall$. So when composing the semantics of the subject `Malta' (:= $\lambda f.\exists x.named(x_a,Malta,ORG) \land f(x)$), the universal will still have wide-scope, as shown below:
\begin{center}
$\forall y.(country(y_a) \to \lnot \exists x.\exists z. named(z_a,  Malta, PER)\land borders(x_ez) \land Theme(x_e, y_a) \land Actor(x_e, z_a))$
\end{center}
\subsection{Negation and universal quantifier} \label{univclass}
\begin{figure}[t]
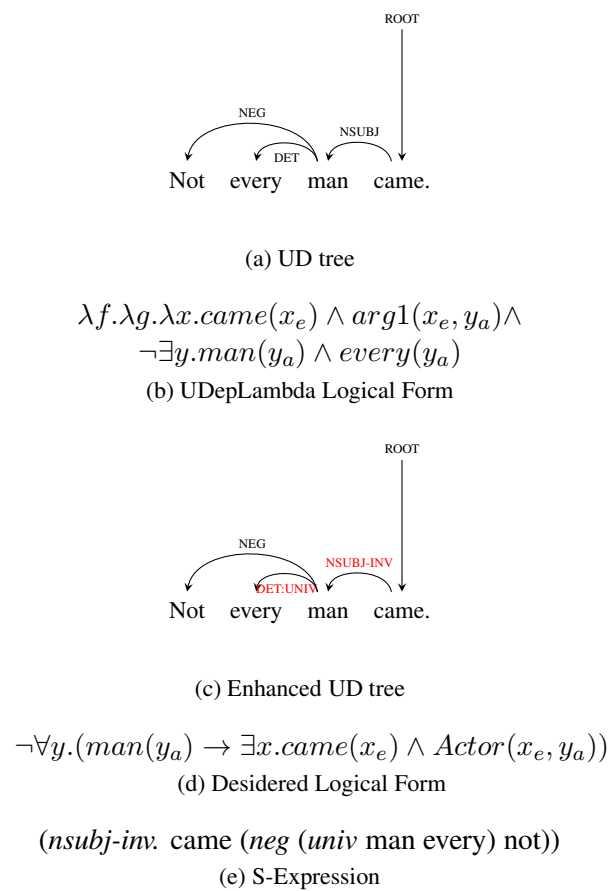

\centering
\subfloat[UD tree]{
\begin{dependency}[theme = simple]
   \begin{deptext}[column sep=0.4em, font=\small]
      Not \& every \& man \& came.\\\\
   \end{deptext}
   \deproot{4}{ROOT}
   \depedge{4}{3}{NSUBJ}
   \depedge[label style={below}]{3}{2}{DET}
   \depedge{3}{1}{NEG}
\end{dependency}
}\\
\subfloat[UDepLambda Logical Form]{
\shortstack{$\lambda f.\lambda g. \lambda x. came(x_e) \land arg1(x_e,y_a) \land$\\$\lnot \exists y. man(y_a) \land every(y_a)$}
}\\
\subfloat[Enhanced UD tree]{
\begin{dependency}[theme = simple]
   \begin{deptext}[column sep=0.4em, font=\small]
      Not \& every \& man \& came.\\\\
   \end{deptext}
   \deproot{4}{ROOT}
   \depedge{4}{3}{\textcolor{red}{NSUBJ-INV}}
   \depedge[label style={below}]{3}{2}{\textcolor{red}{DET:UNIV}}
   \depedge{3}{1}{NEG}
\end{dependency}
}\\
\subfloat[Desidered Logical Form]{
$\lnot \forall y. (man(y_a) \to \exists x. came(x_e) \land Actor(x_e,y_a))$
}\\
\subfloat[S-Expression]{
(\textit{nsubj-inv.} came (\textit{neg} (\textit{univ} man every) not))
}
\caption{The sentence `Not every man came' along with its dependency trees and logical forms}
\label{everynot}
\end{figure}

Alongside quantifiers inherently expressing negation, as the one shown in the previous section, another challenging scope representation arises during the interaction between a negation cue and a non-negative universal quantifier, such `every'. Let's take as example the sentences `Not every man came', shown in Figure \ref{everynot} alongside its FOL representation.\\
\indent If compared to the representation of the sentence `No man came', where the universal quantifier outscopes the negation operator, the construction `not every' yields the opposite interaction where the quantifier is in the scope of $\lnot$ (correspondent to the meaning `there exists some man who came').\\
\indent As shown in the previous section and here in Figure~\ref{everynot}(b), UDepLambda cannot deal with such constructions, yielding a meaning where there exists and event but there doesn't exists the entity that performs it. On the other hand, UDepLambda$\lnot$ can easily derive the correct representation by applying the same edits to the UD graph shown in the previous section. First, we enhance the \textit{det} edge to become a more fine-grained \textit{det:univ} in the presence of the child node `every'. Second, we change \textit{nsubj} into \emph{nsubj-inv.}, since a universal quantifier is in its yield. The lambda expression assigned to the edge \textit{det:univ} is as follows:\\\\
{\small \emph{det:univ}:=$\lambda P.\lambda Q.\lambda f. \forall x.(P(\lambda y. EQ(x,y)) \to f(x))$}\\\\
\indent Once again, we deploy the usual bottom-up binarization hierarchy where all modifiers of a head need to be composed before the head itself can be used for further composition. In the case of `not every...', we start from the modifiers `every' and `not' and compose the edges following the order \textit{det:univ} $\rangle$ \textit{neg} so to make sure that negation operator $\lnot$ outscopes the universal quantifier $\forall$. After the modifiers of the head `man' are composed, we can then move on to compose the head itself with its governor node, the event `came'. The \textit{nsubj:inv.} edge ensures that the subject scopes over the event and not the other way around. Following this, we are able to obtain the final representation:
\begin{center}
$\lnot \forall y. (man(y_a) \to \exists x. came(x_e) \land Actor(x_e,y_a))$
\end{center}

\subsection{Nobody/nothing/nowhere} \label{nonadv}
\begin{figure}
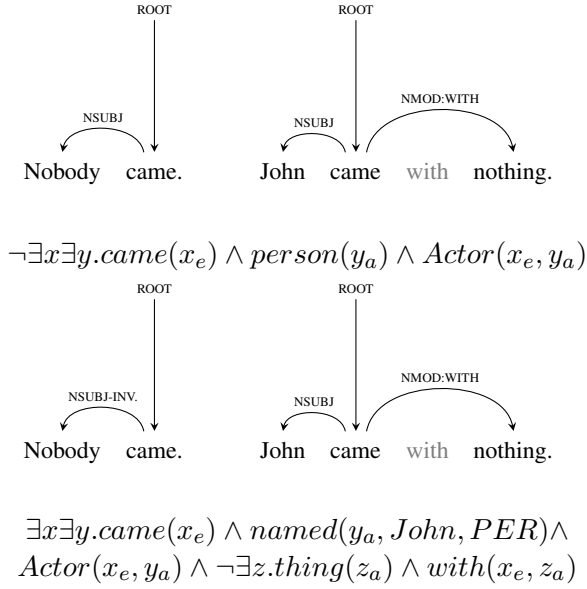

\centering
$\begin{array}{cc}
\begin{dependency}[theme = simple]
   \begin{deptext}[column sep=0.4em,font=\small]
      Nobody \& came.\\\\
   \end{deptext}
   \deproot{2}{ROOT}
   \depedge{2}{1}{NSUBJ}
\end{dependency} &
\begin{dependency}[theme = simple]
   \begin{deptext}[column sep=0.4em, font=\small]
      John \& came \& \textcolor{gray}{with} \& nothing.\\\\
   \end{deptext}
   \deproot{2}{ROOT}
   \depedge{2}{1}{NSUBJ}
   \depedge{2}{4}{NMOD:WITH}
\end{dependency}\\
\multicolumn{2}{c}{
\lnot \exists x \exists y. came(x_e) \land person(y_a) \land Actor(x_e,y_a)
}\\
\begin{dependency}[theme = simple]
   \begin{deptext}[column sep=0.4em,font=\small]
      Nobody \& came.\\\\
   \end{deptext}
   \deproot{2}{ROOT}
   \depedge{2}{1}{NSUBJ-INV.}
\end{dependency} &
\begin{dependency}[theme = simple]
   \begin{deptext}[column sep=0.4em, font=\small]
      John \& came \& \textcolor{gray}{with} \& nothing.\\\\
   \end{deptext}
   \deproot{2}{ROOT}
   \depedge{2}{1}{NSUBJ}
   \depedge{2}{4}{NMOD:WITH}
\end{dependency}\\
\multicolumn{2}{c}{
\shortstack{$\exists x \exists y. came(x_e) \land named(y_a,John,PER) \land $\\$ Actor(x_e,y_a)  \land \lnot \exists z. thing(z_a) \land with(x_e,z_a)$}
}\\
\end{array}$
\caption{Dependency graphs and FOL representations for the sentences `Nobody came' (above) vs. `John came with nothing' (below).}
\label{lexneg}
\end{figure}

\begin{table}
\centering
\begin{tabular}{c|c|c|c}
& nobody & nothing & nowhere\\\hline
nsubj & 7 & 18 & -\\
dobj & - & 34 & -\\
conj & - & 8 & -\\
nsubjpass & 1 & 6 & -\\
root & - & 8 & -\\
advmod & - & - & 3\\
nmod & - & 4 & -\\
other & - & 8 & -\\\hline
tot. & 8 & 86 & 3
\end{tabular}
\caption{Distribution of nobody, nothing and nowhere with their related dependency tags as they appear in the English UD corpus \cite{mcdonald2013universal}}
\label{dist}
\end{table}
As shown in Table \ref{dist}, `nobody', `nothing' and `nowhere' belong to that class of negation cues whose parent edge do not mark them as inherently expressing negation. However using an hand-crafted list of negation cues for English, we can detect and assign them the semantic representation $\lambda f. \lnot \exists x. thing/person/location (x_a) \land f(x)$, where the negation operator scopes over an existentially bound entity.\\
\indent Binarization and composition vary according to whether these elements are arguments or adjuncts. If an argument, the scope of negation includes also the event, otherwise the latter is excluded. To this end, let's compare the sentences `Nobody came' and `John came with nothing', along with their dependency graphs and logic forms (Figure \ref{lexneg}).\\
\indent The argument `nobody' in `Nobody came' yields a scope reading where the negation operator scopes over the existential. To achieve such reading we once again convert the \textit{nsubj} (or any argument edge for that matter) into a \textit{nsubj:inv.} edge. This is reminiscent of how we handled universal quantification when we introduced the quantifier `no', which is in fact integral part of such lexical elements (the semantics of `\textit{no}-body came' can be in fact read as `for all x such that x is a person that x did not come'). Also, the fact that the semantics of these elements is represented through an existential and not a universal bound variable is no problem since we are working under the equivalence $\forall x. P(x) \to \lnot Q(x) \equiv \lnot \exists x. P(x) \land Q(x)$.\\
\indent Given the s-expression (\textit{nsubj:inv.} came nobody) the composition is then as follows:
\begin{center}
$\lnot \exists x.\exists y. person(y_a) \land  f(x) \land Actor(x_e,y_a) \land came(x_e)$
\end{center}
\indent On the other hand, when the negated lexical element is embedded in an adjunct, as in `with nothing', no enhancement of the original dependency edges takes place since we want to preserve negation scope inside the phrase (so to yield a reading where the event `John came' did indeed take place). By substituting and combining the semantics of the s-expression (\textit{nmod:with} came nothing), where the edge \textit{nmod:with} is assigned the lambda expression $\lambda P.\lambda Q.\lambda f.P(\lambda x. f(x) \land Q(\lambda y. with(x_e,y_a)))$, we obtain the following logic form:
\begin{center}
$\lambda f.\exists x. came(x_e) \land f(x) \land \lnot \exists y. (thing(y_a) \land with (x_e,y_a))$
\end{center}
, where we can the scope of negation is limited to the propositional phrase. Given that the $f$ is outside the scope of negation, further compositions (in the case along the edge \textit{nsubj.}) will also compose outside it, yielding the correct form in Figure (\ref{lexneg}).\\
\indent The only limitation we have observed so far concerns `nowhere' (:= $\lambda f. \exists x. location(x_a) \land f(x)$) and the fact it is always associated with a dependency tag \textit{advmod}. The tag \textit{advmod} describes however the manner an action is carried out and has the logical form $\lambda P.\lambda Q.\lambda f. P(\lambda x. f(x) \land Q(\lambda y.Manner(x_e,y_a)))$. This is however different from how `nowhere' is treated in the Groeningen Meaning Bank, where it is described as \emph{where} and not \textit{how} the event takes place. That is, our framework would assign a sentence like `They got nowhere near the money' the logical form $\exists x. got(x_e) \land \lnot \exists y. (location(y_a) \land Manner(x_e,y_a))$, whereas the one contained in the GMB is: $\exists x. got(x_e) \land \lnot \exists y_a. (location(y_a) \land in(x_e,y_a))$

\section{Conclusion and future work}
This paper addressed the problem of representing negation scope from universal dependencies by setting the foundations of \textit{UDepLambda$\lnot$}, a conversion framework whose high-order type theory is able to deal with complex semantic phenomena related to scope. The conversion processes we presented show that it is possible to rely on dependency edges and additionally to minimal language-dependent lexical information to compose the semantics of negation scope. The fact that this formalism is able to correctly compose the scope for many complex phenomena related to negation scope is promising.\\
\indent We are currently working on extending this work in two directions:\\
{\bf 1.} \textit{Automatic framework evaluation}: given the conversion rules presented in this paper, we are planning to automatically convert the UD graphs for the sentences in the GMB so to compare the graph we automatically generate with a gold-standard representation. This would also to identify and quantify the errors of our framework.\\
{\bf 2.} \textit{Automatic semantic parsing}: given the connection between this framework and the GMB, we would like to explore the possibility of learning the conversion automatically, so not to rely on an hand-crafted hierarchy to decide the order of edge traversal.

\bibliographystyle{eacl2017}
\bibliography{eacl2017}

\newpage
\onecolumn

\appendix
\section{Step-by-step $\lambda$-reductions}
*Throughout the derivations, we are going to use the variable $e$ in place of $x_e$ and $z$,$y$ or $x$ in place of $x_a$. Due to space restrictions, we skip reduction for existential closure ($\to_{ex-clos}$).
\subsection{`John does not eat'}
\begin{math}
\lambda P.\lambda Q.\lambda f. P(\lambda e. f(e) \land Q(\lambda x.Actor(e,x))) (\lambda f.\exists e. eat(e) \land f(e))\\
\to_\alpha \lambda P.\lambda Q.\lambda f. P(\lambda e. f(e) \land Q(\lambda x.Actor(e,x))) (\lambda g.\exists e'. eat(e') \land g(e'))\\
\to_\beta \lambda Q.\lambda f.\lambda g.[\exists e'. eat(e') \land g(e')] (\lambda e. f(e) \land Q(\lambda x.Actor(e,x)))\\
\to_\beta \lambda Q.\lambda f. \exists e'. eat(e') \land \lambda e.[f(e) \land Q (\lambda x.Actor(e,x))](e')\\
\to_\beta \lambda Q.\lambda f. \exists e'. eat(e') \land f(e') \land Q (\lambda x.Actor(e',x))\\
\to_\beta \lambda Q.\lambda f. \exists e'. eat(e') \land f(e') \land Q[\lambda x.Actor(e',x)](\lambda f.\exists x. named(x,John,PER) \land f(x))\\
\to_\alpha \lambda Q.\lambda f. \exists e'. eat(e') \land f(e') \land Q[\lambda x.Actor(e',x)](\lambda g.\exists z. named(z,John,PER) \land g(z))\\
\to_\beta \lambda f. \exists e'. eat(e')  \land f(e') \land \lambda g.[\exists z. named(z,John,PER) \land g(z)](\lambda x.Actor(e',x))\\
\to_\beta \lambda f. \exists e'. eat(e')  \land f(e') \land \exists z. named(z,John,PER) \land \lambda x.[Actor(e',x)](z)\\
\to_\beta \lambda f. \exists e'. eat(e') \land f(e') \land \exists z. named(z,John,PER) \land Actor(e',z)\\\\
\lambda P.\lambda Q.\lambda f. \lnot P(\lambda e.f(e)) (\lambda f. \exists e'. \exists z. eat(e') \land f(e') \land named(z,John,PER) \land Actor(e',z))\\
\to_\alpha \lambda P.\lambda Q.\lambda f. \lnot P(\lambda e.f(e))(\lambda g. \exists e'. \exists z. eat(e') \land g(e') \land named(z,John,PER) \land Actor(e',z))\\
\to_\beta \lambda Q.\lambda f.\lnot \lambda g. [\exists e'. \exists z. eat(e') \land g(e') \land named(z,John,PER) \land Actor(e',z)] (\lambda e.f(e))\\
\to_\beta \lambda Q.\lambda f. \lnot \exists e'. \exists z. eat(e') \land \lambda e.[f(e)](e') \land named(z,John,PER) \land Actor(e',z)\\
\to_\beta \lambda Q.\lambda f. \lnot \exists e'. \exists z. eat(e') \land f(e') \land named(z,John,PER) \land Actor(e',z)\\
\to_\beta \lambda f. \lnot \exists e'. \exists z. eat(e') \land f(e') \land named(z,John,PER) \land Actor(e',z)\\
\to_\beta \lambda f.[\lnot \exists e'. \exists z. eat(e') \land named(z,John,PER) \land Actor(e',z) \land f(e')](\lambda x.TRUE)\\
\to_\beta \lnot \exists e'. \exists z. eat(e') \land named(z,John,PER) \land Actor(e',z) \land \lambda x.[TRUE](e')\\
\to_\beta {\bf \lnot \exists e'. \exists z. eat(e') \land named(z,John,PER) \land Actor(e',z)}
\end{math}

\subsection{`No man came'}
$\lambda P.\lambda Q.\lambda f.\forall x.(P(\lambda y. EQ(x,y)) \to \lnot f(x))(\lambda f. \exists x.man(x) \land f(x))$\\          
$\to_\alpha \lambda P.\lambda Q.\lambda f. \forall x.(P(\lambda y. EQ(x,y)) \to \lnot f(x))(\lambda f'.\exists z.man(z) \land f'(z))$\\
$\to_\beta \lambda Q.\lambda f. \forall x. (\lambda f'.[\exists z.man(z) \land f'(z)](\lambda y. EQ(x,y)) \to \lnot f(x))$\\
$\to_\beta \lambda Q.\lambda f.\forall x.(\exists z.man(z) \land \lambda y.[EQ(x,y)](z) \to \lnot f(x))$\\
$\to_\beta \lambda Q.\lambda f.\forall x.(\exists z.man(z) \land EQ(x,z) \to \lnot f(x))$\\
$\to_{EQ} \lambda Q.\lambda f.\forall x.(man(x) \to \lnot f(x))$\\
$\to_\beta \lambda f. \forall x.(man(x) \to \lnot f(x))$\\\\
$\lambda P.\lambda Q.\lambda f. Q(\lambda x. P(\lambda e.Actor(e,x) \land f(e)))(\lambda f.\exists e. came(e) \land f(e))$\\
$\to_\alpha \lambda P.\lambda Q.\lambda f. Q(\lambda x. P(\lambda e.Actor(e,x) \land f(e)))(\lambda g.\exists e'. came(e') \land g(e'))$\\
$\to_\beta \lambda Q.\lambda f. Q(\lambda x. \lambda g.[\exists e'. came(e') \land g (e'))](\lambda e.Actor(e,x) \land f(e)))$\\
$\to_\beta \lambda Q.\lambda f. Q(\lambda x.\exists e'. came(e') \land \lambda e.[Actor(e,x) \land f(e)](e'))$\\
$\to_\beta \lambda Q.\lambda f. Q(\lambda x.\exists e'. came(e') \land Actor(e',x) \land f(e'))$\\
$\to_\alpha \lambda Q.\lambda f. Q(\lambda x.\exists e'. came(e') \land Actor(e',x) \land f(e'))(\lambda f'. \forall x'.(man(x') \to \lnot f'(x'))$\\
$\to_\beta \lambda f. \lambda f'.[\forall x'.(man(x') \to \lnot f'(x'))](\lambda x.\exists e'. came(e') \land Actor(e',x) \land f(e'))$\\
$\to_\beta \lambda f.\forall x'.(man(x') \to \lnot\lambda x.[\exists e'. came(e') \land Actor(e',x) \land f(e')](x'))$\\
$\to_\beta \lambda f.\forall x'.(man(x') \to \lnot \exists e'. came(e')\land Actor(e',x') \land f(e'))$\\
$\to_{ex-clos.}{\bf \forall x'.(man(x') \to \lnot \exists e'. came(e') \land Actor(e',x'))}$

\subsection{`Not every man came'}
% $\lambda P.\lambda Q.\lambda f.\forall x.(P(\lambda y. EQ(x,y)) \to f(x))(\lambda f. \exists x.man(x) \land f(x))$\\          
% $\to_\alpha \lambda P.\lambda Q.\lambda f. \forall x.(P(\lambda y. EQ(x,y)) \to f(x))(\lambda f'.\exists z.man(z) \land f'(z))$\\
% $\to_\beta \lambda Q.\lambda f. \forall x. (\lambda f'.[\exists z.man(z) \land f'(z)](\lambda y. EQ(x,y)) \to f(x))$\\
% $\to_\beta \lambda Q.\lambda f.\forall x.(\exists z.man(z) \land \lambda y.[EQ(x,y)](z) \to f(x))$\\
% $\to_\beta \lambda Q.\lambda f.\forall x.(\exists z.man(z) \land EQ(x,z) \to f(x))$\\
% $\to_{EQ} \lambda Q.\lambda f.\forall x.(man(x) \to f(x))$\\
$\to_\forall \lambda f. \forall x.(man(x) \to f(x))$\\
% $\lambda P.\lambda Q.\lambda f. \lnot P(\lambda x.f(x))(\lambda f. \forall x.(man(x) \to f(x)))$\\
% $\to_\alpha \lambda P.\lambda Q.\lambda f. \lnot P(\lambda x.f(x))(\lambda g. \forall z.(man(z) \to g(z)))$\\
% $ \to_\beta \lambda Q.\lambda f. \lnot \lambda g.[\forall z.(man(z) \to g(z))](\lambda x.f(x))$\\
% $ \to_\beta \lambda Q.\lambda f. \lnot \forall z.(man(z) \to \lambda x.[f(x)](z))$\\
% $ \to_\beta \lambda Q.\lambda f. \lnot \forall z.(man(z) \to f(z))$\\
$ \to_\neg \lambda f. \lnot \forall z.(man(z) \to f(z))$\\\\
$\lambda P.\lambda Q.\lambda f. Q(\lambda x. P(\lambda e.Actor(e,x) \land f(e)))(\lambda f.\exists e. came(e) \land f(e))$\\
$\to_\alpha \lambda P.\lambda Q.\lambda f. Q(\lambda x. P(\lambda e.Actor(e,x) \land f(e)))(\lambda g.\exists e'. came(e') \land g(e'))$\\
$\to_\beta \lambda Q.\lambda f. Q(\lambda x. \lambda g.[\exists e'. came(e') \land g (e'))](\lambda e.Actor(e,x) \land f(e)))$\\
$\to_\beta \lambda Q.\lambda f. Q(\lambda x.\exists e'. came(e') \land \lambda e.[Actor(e,x) \land f(e)](e'))$\\
$\to_\beta \lambda Q.\lambda f. Q(\lambda x.\exists e'. came(e') \land Actor(e',x) \land f(e'))$\\\\
$\lambda Q.\lambda f. Q(\lambda x.\exists e'. came(e') \land Actor(e',x) \land f(e'))(\lambda f. \lnot \forall z.(man(z) \to f(z))$\\
$\to_\alpha \lambda Q.\lambda f. Q(\lambda x.\exists e'. came(e') \land Actor(e',x) \land f(e'))(\lambda f'. \lnot \forall z.(man(z) \to f'(z))$\\
$\to_\beta \lambda f. \lambda f'.[\lnot \forall z.(man(z) \to f'(z))](\lambda x.\exists e'. came(e') \land Actor(e',x) \land f(e'))$\\
$\to_\beta \lambda f. \lnot \forall z.(man(z) \to \lambda x.[\exists e'. came(e') \land Actor(e',x) \land f(e')](z))$\\
$\to_\beta \lambda f. \lnot \forall z.(man(z) \to \exists e'. came(e')\land Actor(e',z) \land f(e'))$\\
$\to_{ex-clos.} {\bf \lnot \forall z.(man(z) \to \exists e'. came(e')\land Actor(e',z))}$

\subsection{`Nobody came'}
$\lambda P.\lambda Q.\lambda f.Q(\lambda x. P(\lambda e.f(e) \land Actor(e,x)))(\lambda f. \exists e. f(e) \land came(e))$\\
$\to_\alpha \lambda P.\lambda Q.\lambda f.Q(\lambda x. P(\lambda e.f(e) \land Actor(e,x)))(\lambda g. \exists e'. g(e') \land came(e'))$\\
$\to_\beta \lambda Q.\lambda f.Q(\lambda x. \lambda g.[\exists e'.g(e') \land came(e')](\lambda e.f(e) \land Actor(e,x)))$\\
$\to_\beta \lambda Q.\lambda f.Q(\lambda x. \exists e'.\lambda e.[f(e) \land Actor(e,x)](e') \land came(e'))$\\
$\to_\beta \lambda Q.\lambda f.Q(\lambda x. \exists e'. f(e') \land Actor(e',x) \land came(e'))$\\\\
$\lambda Q.\lambda f.Q(\lambda x. \exists e'. f(e') \land Actor(e',x) \land came(e'))(\lambda f.\lnot \exists x. person(x) \land f(x))$\\
$\to_\alpha \lambda Q.\lambda f.Q(\lambda x. \exists e'. f(e') \land Actor(e',x) \land came(e'))(\lambda g.\lnot \exists z. person(z) \land g(z))$\\
$\to_\beta \lambda f.\lambda g.[\lnot \exists z. person(z) \land g(z)](\lambda x. \exists e'. f(e') \land Actor(e',x) \land came(e'))$\\
$\to_\beta \lambda f.\lnot \exists z. person(z) \land \lambda x. [\exists e'. f(e') \land Actor(e',x) \land came(e')](z)$\\
$\to_\beta \lambda f.\lnot \exists z. \exists e'. person(z) \land  f(e') \land Actor(e',z) \land came(e')$\\
$\to_{ex-clos.}{\bf \lnot \exists z. \exists e'. person(z) \land Actor(e',z) \land came(e')}$

\subsection{`John came with nothing'}
$\lambda P.\lambda Q.\lambda f. P(\lambda e. f(e) \land Q(\lambda x.with(e,x))) (\lambda f.\exists e. came(e) \land f(e))$\\
$\to_\alpha \lambda P.\lambda Q.\lambda f. P(\lambda e. f(e) \land Q(\lambda x.with(e,x))) (\lambda g.\exists e'. came(e') \land g(e'))$\\
$\to_\beta \lambda Q.\lambda f.\lambda g.[\exists e'. came(e') \land g(e')] (\lambda e. f(e) \land Q (\lambda x.with(e,x)))$\\
$\to_\beta \lambda Q.\lambda f. \exists e'. came(e') \land f(e') \land \lambda e.[Q (\lambda x.with(e,x))](e')$\\
$\to_\beta \lambda Q.\lambda f. \exists e'. came(e')  \land f(e') \land Q (\lambda x.with(e',x))$\\
$\to_\beta \lambda Q.\lambda f. \exists e'. came(e') \land f(e') \land Q[\lambda x.with(e',x)](\lambda f.\lnot \exists x. thing(x) \land f(x))$\\
$\to_\alpha \lambda Q.\lambda f. \exists e'. came(e')  \land f(e') \land Q[\lambda x.with(e',x)](\lambda g.\lnot \exists z. thing(z) \land g(z))$\\
$\to_\beta \lambda f. \exists e'. came(e')  \land f(e') \land \lambda g.[\lnot \exists z. thing(z) \land g(z)](\lambda x.with(e',x))$\\
$\to_\beta \lambda f. \exists e'. came(e') \land f(e') \land \lnot \exists z. thing(z) \land \lambda x.[with(e',x)](z)$\\
$\to_\beta \lambda f. \exists e'. came(e') \land f(e') \land \lnot \exists z. thing(z) \land with(e',z)$\\\\
$\lambda P. \lambda Q. \lambda f. P(\lambda e. f(e) \land Q(\lambda x. Actor(e,x)))(\lambda f. \exists e'. came(e') \land f(e') \land \lnot \exists z. thing(z) \land with(e',z))$\\
$\to_\alpha \lambda P. \lambda Q. \lambda f. P(\lambda e. f(e) \land Q(\lambda x. Actor(e,x)))(\lambda g. \exists e'. came(e') \land g(e') \land \lnot \exists z. thing(z) \land with(e',z))$\\
$\to_\beta \lambda Q. \lambda f. \lambda g. [\exists e'. came(e') \land g(e') \land \lnot \exists z. thing(z) \land with(e',z)](\lambda e. f(e) \land  Q(\lambda x. Actor(e,x)))$\\
$\to_\beta \lambda Q. \lambda f. \exists e'. came(e') \land \lambda e.[f(e) \land  Q(\lambda x. Actor(e,x))](e') \land \lnot \exists z. thing(z) \land with(e',z)$\\
$\to_\beta \lambda Q. [\lambda f. \exists e'. came(e') \land  f(e') \land Q(\lambda x.Actor(e',x)) \land \lnot \exists z. thing(z) \land with(e',z)]$\\$(\lambda g. \exists y. named(y,John,PER) \land g(y))$\\
$\to_\beta \lambda f. \exists e'. came(e') \land f(e') \land \lambda g. [\exists y. named(y,John,PER)\land g(y)](\lambda x. Actor(e',x)) \land \lnot \exists z. thing(z) \land with(e',z)]$\\
$\to_\beta \lambda f. \exists e'. came(e') \land f(e') \land \exists y. named(y,John,PER) \land \lambda x. [Actor(e',x)](y) \land \lnot \exists z. thing(z) \land with(e',z)$\\
$\to_\beta \lambda f. \exists e'. came(e') \land \exists y. named(y,John,PER) \land f(e') \land Actor(e',y) \land \lnot \exists z. thing(z) \land with(e',z)$\\
$\to_{ex-clos.} {\bf \exists e'.\exists y. came(e') \land named(y,John,PER) \land Actor(e',y) \land \lnot \exists z. thing(z) \land with(e',z)}$\\

\end{document}